\documentclass[10pt,twocolumn,letterpaper]{article}

\usepackage{cvpr}              % To produce the CAMERA-READY version
\usepackage{graphicx}
\usepackage{amsmath}
\usepackage{amssymb}
\usepackage{booktabs}

 \usepackage{color}
\usepackage[pagebackref,breaklinks,colorlinks]{hyperref}

\usepackage[capitalize]{cleveref}
\crefname{section}{Sec.}{Secs.}
\Crefname{section}{Section}{Sections}
\Crefname{table}{Table}{Tables}
\crefname{table}{Tab.}{Tabs.}

\graphicspath{figures/}

\def\cvprPaperID{13} % *** Enter the CVPR Paper ID here
\def\confName{CVPR}
\def\confYear{2022}

\begin{document}

\title{Leaf Tar Spot Detection Using RGB Images}

\author{\parbox{16cm}{\centering
    {Sriram Baireddy$^{\star}$ \quad Da-Young Lee$^{\dagger}$ \\ Carlos Gongora-Canul$^{\dagger, \ddagger}$ \quad Christian D. Cruz$^{\dagger}$ \quad Edward J. Delp$^{\star}$}\\
    {\normalsize
    $^{\star}$ Video and Image Processing Lab (VIPER), Purdue University, West Lafayette, Indiana, USA\\
    $^{\dagger}$ Department of Botany and Plant Pathology, Purdue University, West Lafayette, Indiana, USA\\
    $^{\ddagger}$ Tecnologico Nacional de Mexico, Instituto Tecnologico de Conkal, Mexico
    }
}}
\maketitle

%%%%%%%%% ABSTRACT
\begin{abstract}
  Tar spot disease is a fungal disease that appears as a series of black circular spots containing spores on corn leaves.
  Tar spot has proven to be an impactful disease in terms of reducing crop yield.
  To quantify disease progression, experts usually have to visually phenotype leaves from the plant.
  This process is very time-consuming and is difficult to incorporate in any high-throughput phenotyping system.
  Deep neural networks could provide quick, automated tar spot detection with sufficient ground truth.
  However, manually labeling tar spots in images to serve as ground truth is also tedious and time-consuming.
  In this paper we first describe an approach that uses automated image analysis tools to generate ground truth images that are then used for training a Mask R-CNN.
  We show that a Mask R-CNN can be used effectively to detect tar spots in close-up images of leaf surfaces.
  We additionally show that the Mask R-CNN can also be used for in-field images of whole leaves to capture the number of tar spots and area of the leaf infected by the disease.
\end{abstract}

%%%%%%%%% BODY TEXT
\section{Introduction}

Tar spot disease is a fungal disease that appears as a series of black circular spots containing spores on 
corn leaves~\cite{HKR1995}.
It has proven to be an impactful disease in terms of reducing yield in affected crop fields~\cite{bajet_1994}. 
Economic damage up to 50\% has been documented in Latin America when epidemics are severe early in corn plants' reproductive phases~\cite{cruz_2020}.
Tar spot detection has been done using human visual disease assessments of `stromata', which are black circular structures produced by the tar spot pathogen \textit{Phyllachora maydis}.
In general, these stromata start off circular, but can quickly elongate and merge as the disease progresses. 
To capture disease progression, experts in plant pathology quantify disease intensity using visual methods, such as a manual count of the tar spots.
In addition to count, the relative area of the leaf covered in tar spots is a good measure of the disease severity~\cite{nutter_1997, nutter_2001, lee_2021}.
This manual analysis can be done using computer-based tools using an image of the leaf, or physically, but either way, the process is very time consuming.

Despite the alarming nature of tar spot disease, the current capabilities available make it difficult to do any sort of high-throughput phenotyping at the field level.
Thus, we aim to automate this process and reduce the required overhead using image processing techniques and deep learning.

\begin{figure}[tbp]
   \begin{center}
   \includegraphics[width=\linewidth]{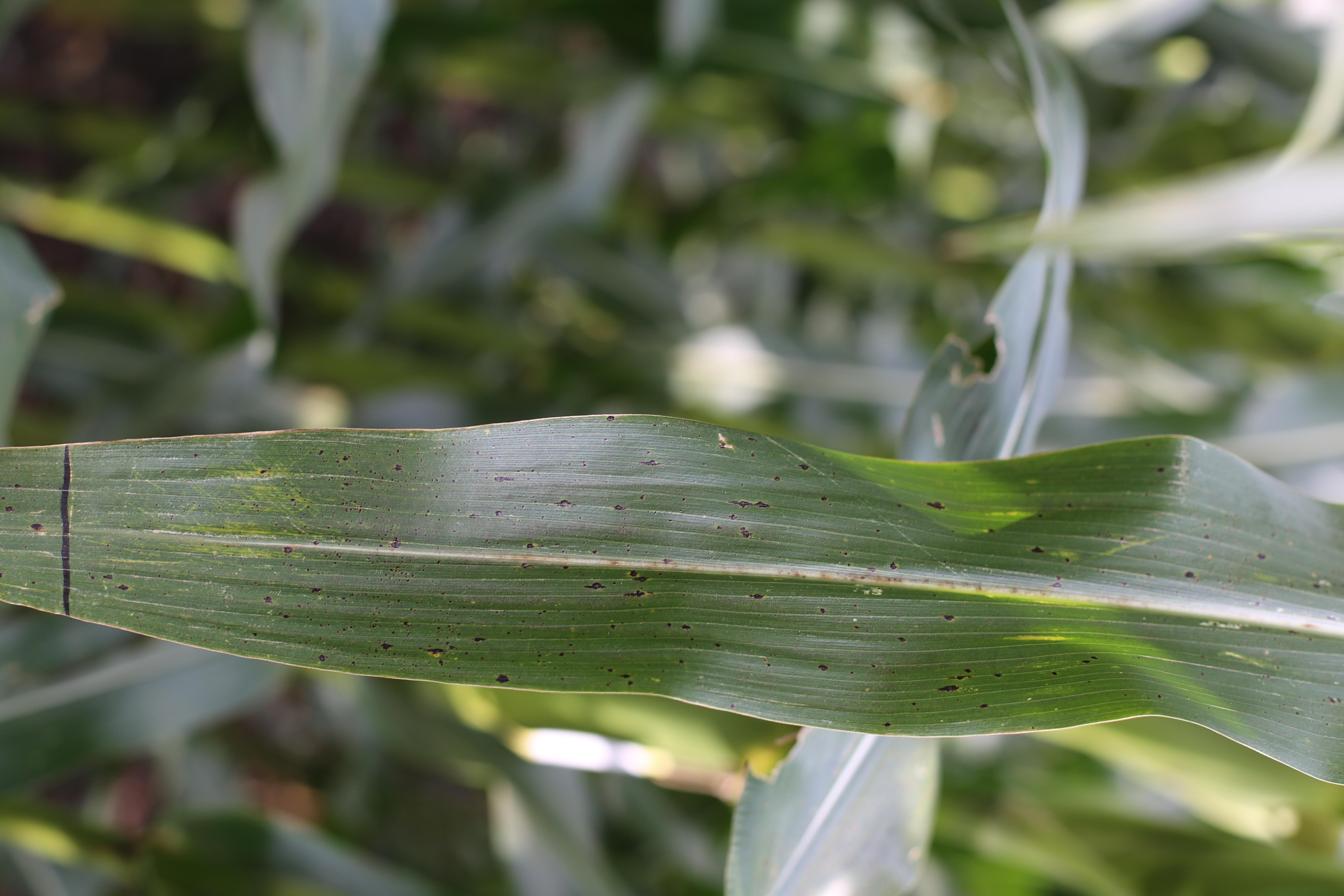}
   \end{center}
      \caption{An example of a leaf showing signs of tar spot disease. There are numerous black spots covering the surface of the leaf.}
   \label{fig:example_tar_spot}
\end{figure}

Deep learning is a rapidly growing field that has progressed quickly in the last decade. 
Neural networks have been shown to be very effective at several computer vision and image processing tasks such as object detection~\cite{zhao_2019} and image segmentation~\cite{minaee_2020}, making them an ideal tool to use in detecting tar spots.
Neural networks  can leverage the parallel processing and computational power of GPUs to process large amounts of image data rapidly, making them ideal for high-throughput phenotyping.
However, neural network models require large amounts of training data to achieve optimal performance.
As discussed, generating annotated examples of leaf tar spots is a time consuming process that requires manual labeling  from  subject matter experts.
In this paper, we show that a convolutional neural network (CNN), specifically Mask R-CNN, can be trained using ``ground truth'' data generated from images using automated image analysis techniques.
To be specific, given an RGB image of a leaf of a plant with tar spot, we first designed a traditional image analysis method using techniques such as thresholding and connected components to identify and segment tar spots.
The idea here is that we use an automated approach to generate ``ground truth'' tar spot masks that we can use for
 training our neural network. 
We assume the masks we obtain to be accurate ground truth viable for training.
This process involves no manual labeling by expert plant pathologists, so the amount of resources that need to be invested is reduced significantly.
%\textbf{{\color{red} This statement ``ground truth data generated from images using image processing techniques'' is confuseing to me. What do you actually mean here? Explain the process in more detail here.Does the process invovle any manual labeling?}}
Our Mask R-CNN model, trained on these ground truth masks, achieves reasonable performance on our limited expert-labeled dataset, and can be used to process in-field images of leaves with tar spot.

\section{Related Work}
\subsection{Image Based Plant Phenotyping}
Manual examination of plants has been fairly effective at evaluating the health and growth of the plant.
However, in recent years, with the sheer volume of field crops grown, manual inspection is steadily becoming infeasible, and the accuracy of visual disease assessments is being questioned due to the ``human factor''~\cite{nutter_2006}.
Using images adds scalability to the analysis conducted, and also introduces a factor of replicability and objectivity when quantifying any plant phenotype.

Image-based plant phenotyping can be categorized into two approaches based on the environment in which the images are acquired~\cite{CSA2019,choudury_2016}.
%\textbf{{\color{red} You need to put some reference cites in here}}
The first is under controlled conditions i.e., the imaging environment is consistent and the plants are grown with minimal unexpected interference.
For example, the indoor high-throughput plant phenotyping system proposed in \cite{falhgren_2015} allows plants to be freely moved around to be watered, weighted, and imaged.
These environments have been shown to be successful at enabling high precision in trait estimation and growth projection~\cite{falhgren_2015}.
In \cite{lu_2010}, leaves are destructively phenotyped to obtain the leaf area measurement from the number of pixels in the segmentation mask and known physical dimensions.
In \cite{minervini_2014}, plants in an automated indoor facility are detected using a variety of image processing techniques like color thresholding, k-means, and active contours.
While phenotyping in a controlled environment is useful, it cannot be extended to large scale field studies, which are conducted outdoors.

The other approach to image-based plant phenotyping is estimating plant traits using images collected in an outdoor area.
These images tend to have more background variations and are not as uniform.
Data collection for outdoor plant phenotyping can be done in several ways.
In \cite{liu_2010}, field images collected using a hand-held device are segmented using histogram thresholding.
The resulting segmentation masks are then used to estimate leaf surface area per unit area of land.
For the specific task of quantifying tar spot disease progression, in \cite{oh_2021}, the authors proposed using a combination of regression methods and observable characteristics from overhead images captured by unmanned aerial vehicles (UAVs).
Characteristics such as canopy cover and volume were extracted from these images and used to fit regression models like support vector regression to visual estimations of disease severity performed by expert plant pathologists.
Another approach~\cite{lee_2021} used image processing techniques to create a contour-based tar spot detection approach that leads to promising results in tar spot detection and area estimation, but this approach lacks automation.

In the work presented in this paper, we use images collected in an outdoor area.
The dataset consists of both close-up images of leaves with tar spot, as well as images of the whole leaf that are better suited for quantifying disease progression in the crop fields.
Visual examination of these leaves by experts is difficult and time-consuming, and so we automate the process using image processing and deep learning, which will be discussed in Section \ref{sec:approach}.
%\textbf{{\color{red} Finish the sentence and say more}}

\subsection{Object Detection and Image Segmentation}
%Great strides have been made to improve the performance of neural networks in various tasks to a superhuman level. 
Let us consider object recognition, which is a general term for computer vision tasks that involve identifying objects in digital images. 
Image classification is the act of labeling an entire image with the class of one of its contained objects i.e., an image of cat being labeled ‘cat’. 
Another task, object localization, involves finding the location of the objects in an image and drawing a bounding box around them. 
Object detection, is a combination of these tasks, where all the objects in an image are found and classified (or labeled). 
An image of three different animals would have three different bounding boxes and class labels.

\begin{figure}[bp]
   \begin{center}
   \includegraphics[width=\linewidth]{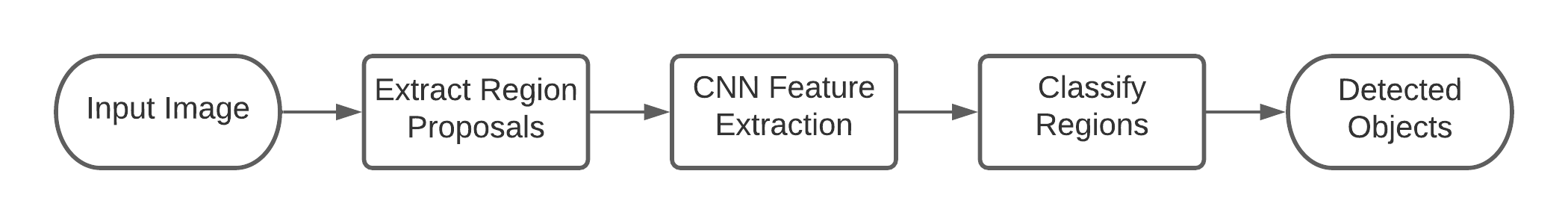}
   \end{center}
      \caption{Block diagram showing the general R-CNNs process.}
   \label{fig:basic_rcnn}
\end{figure}

One approach to object detection using neural networks is the R-CNN~\cite{rcnn}, or Region-based Convolutional Neural Networks, family of models. 
The general idea behind these models is shown in Figure \ref{fig:basic_rcnn}, consisting of three stages. Given an input image, the first step is to identify class independent areas of interest i.e., mark possibly interesting bounding boxes. 
Then the areas within these bounding boxes are passed through a CNN to extract important features. 
These features are then used to assign the bounding boxes one of the known labels. 
Initially this approach consisted of three separate networks (one for each stage). 
Fast R-CNN~\cite{fast-rcnn} is a single model that speeds up the entire learning process by using spatial pyramid pooling networks, or SPPnets~\cite{sppnet}. 
Additional optimization was made when Faster R-CNN~\cite{faster-rcnn} was introduced, which used a Region Proposal Network (RPN) to refine region proposals as part of the training process. 
Effective image segmentation can be achieved by using additional convolutional layers in parallel to generate object masks, as seen with Mask R-CNN~\cite{mask-rcnn}.

\section{Our Approach}
\label{sec:approach}

\subsection{Generating Ground Truth Tar Spot Images}
Most deep learning object detection approaches require labeled data, which is a tedious and time-consuming process of manual work from experts. 
We instead decided to train our tar spot detection network with labeled data generated using automated image analysis techniques that exploits properties of a typical tar spot image (e.g., the dark tar spots).
%\textbf{{\color{red} This statement ``data generated from images using image processing techniques'' is confusing to me. What do you actually mean here? More detail is needed}}
To be specific, given an RGB image of a leaf of a plant infected with tar spot, we first designed a traditional image analysis method using techniques such as thresholding and connected components to identify and segment tar spots, a process we call ``automatic ground truthing.''
The idea here is that we use an automated approach to generate ground truth tar spot masks that we can use for training. 
We assume the masks we obtain to be accurate ground truth viable for training.
This process involves no manual labeling by expert plant pathologists, so the amount of resources that need to be invested is reduced significantly.
Note that we have a limited set of ground truth images labeled by plant experts that we will use for testing the performance of method trained using this `automatic ground truthing'' approach.

\begin{figure}[tbp]
   \begin{center}
   \includegraphics[width=\linewidth]{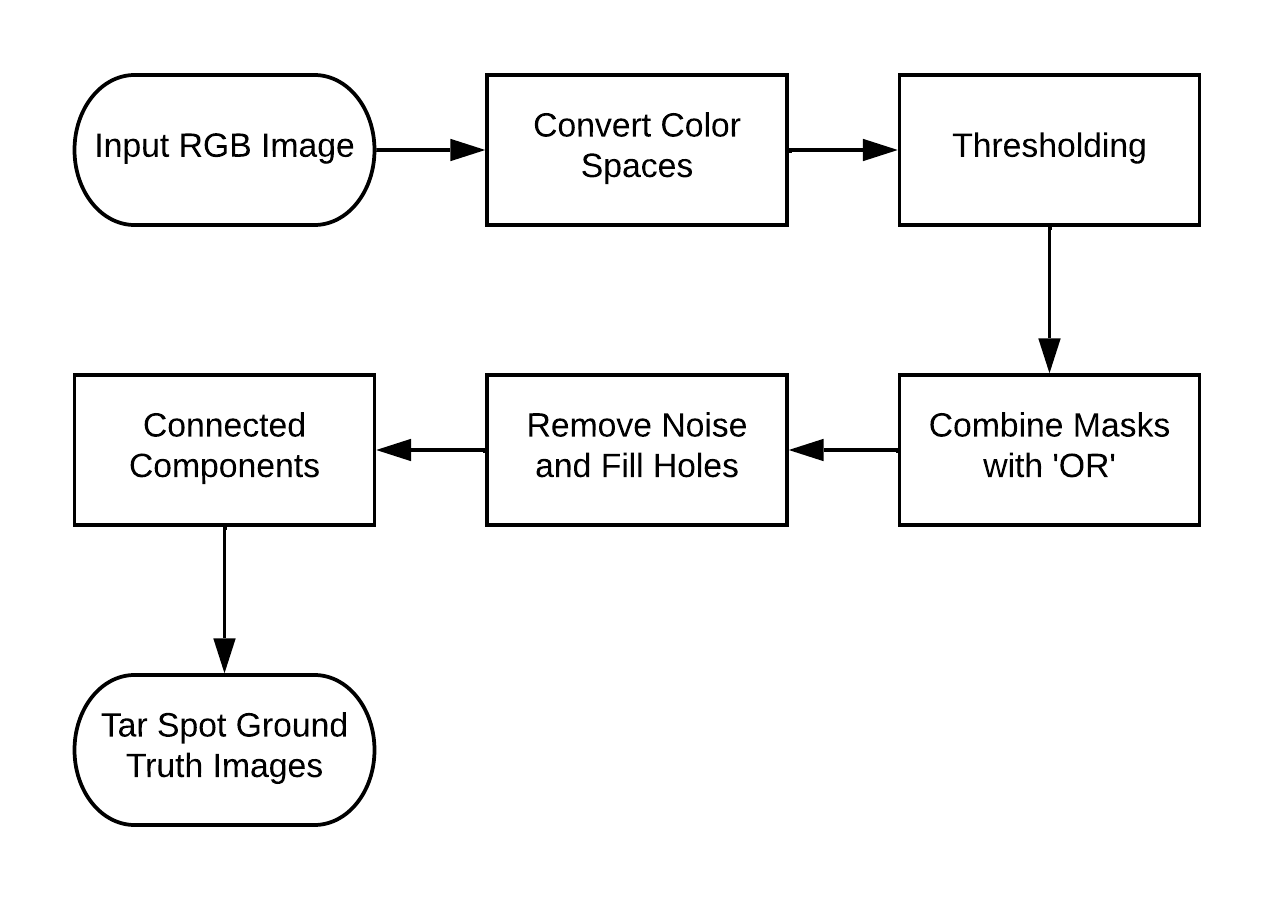}
   \end{center}
      \caption{A block diagram of our automatic ground truthing approach for tar spot images. }
   \label{fig:our_approach}
\end{figure}

The block diagram of our automatic ground truthing approach is shown in Figure \ref{fig:our_approach}. 
We convert the RGB image to both the HSV and L*A*B* color spaces. 
We then threshold the V channel in the HSV space and the A* channel in the L*A*B* color space using thresholds determined empirically.
This process provides us with two preliminary tar spot masks. 
We combine these masks using the logical ‘OR’ operation into a single tar spot mask containing information from the V and A* channels. 
To remove any undesired objects in the mask, we use sequential opening and closing operations~\cite{Har1987,morphology} with a 3x3 structuring element. 
This removes noise and fills holes in the mask. 
The final step is to use connected component labeling~\cite{connected-components} on the mask to separate out each instance of a tar spot. 
These results are then used to train the Mask R-CNN.

\subsection{Mask R-CNN}
Mask R-CNN~\cite{mask-rcnn} is an extension of the object detection model Faster R-CNN~\cite{faster-rcnn}.
Faster R-CNN consists of two stages.
The first stage is known as a Region Proposal Network (RPN), which proposes potential object bounding boxes.
The second stage is essentially a Fast R-CNN~\cite{fast-rcnn}, since it extracts features from these potential object bounding boxes to assign labels and refine the bounding boxes.
Mask R-CNN~\cite{mask-rcnn} uses the same two-stage procedure as Faster R-CNN~\cite{faster-rcnn}.
The first stage consisting of the RPN is kept the same.
The second stage, in parallel to learning the classification and refining the bounding boxes, also generates a binary mask output for each region of interest.
The novelty here is that the mask output and the classification are done in parallel, as opposed to the classification depending on the mask~\cite{mask-rcnn}.
Mask R-CNN has proven to be a very good approach for object detection, which can be partially attributed to the impact of multi-task learning~\cite{multi-task-learning}. 
Due to the process of learning the general location of objects (via bounding boxes), types of objects (via class labels), and specific location of objects (via object masks), Mask R-CNN learns more related information and connections between its various inputs. 
We use a Mask R-CNN for tar spot detection task due to its high level of performance.

\section{Experimental Results}

%\textbf{{\color{red} Has there been any other work published in tar spot detection? I am concerned that the reviewers might complain that we did not compare our method to anything else.Do we cite any other tar spot detection papers?}}
%\textbf{\color{blue} I have added some more in the related work section.}

%\begin{figure*}[tbp]
%   \begin{center}
%   \includegraphics[width=0.9\linewidth]{figures/mask_rcnn_output.png}
%   \end{center}
%      \caption{Tar spots detected by our trained Mask R-CNN. Note that the image also shows the presence of plant disease symptoms other than tar spot, in addition to small debris often found on corn leaves in the field.}
%   \label{fig:orig_results}
%\end{figure*}

\begin{figure*}[t]
   \centering
   \begin{subfigure}[t]{0.33\textwidth}
       \includegraphics[width=\textwidth]
       {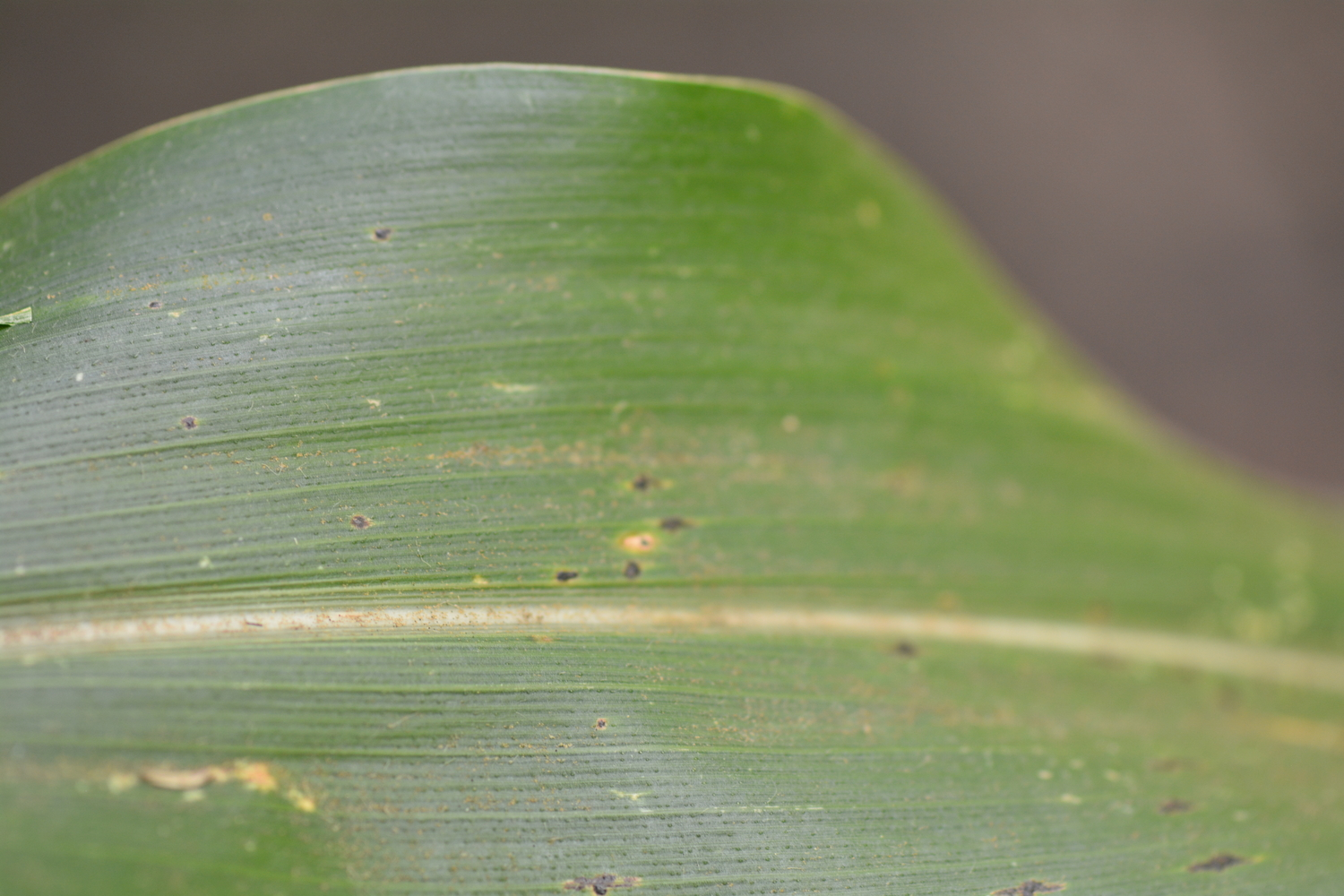}
       \includegraphics[width=\textwidth]
       {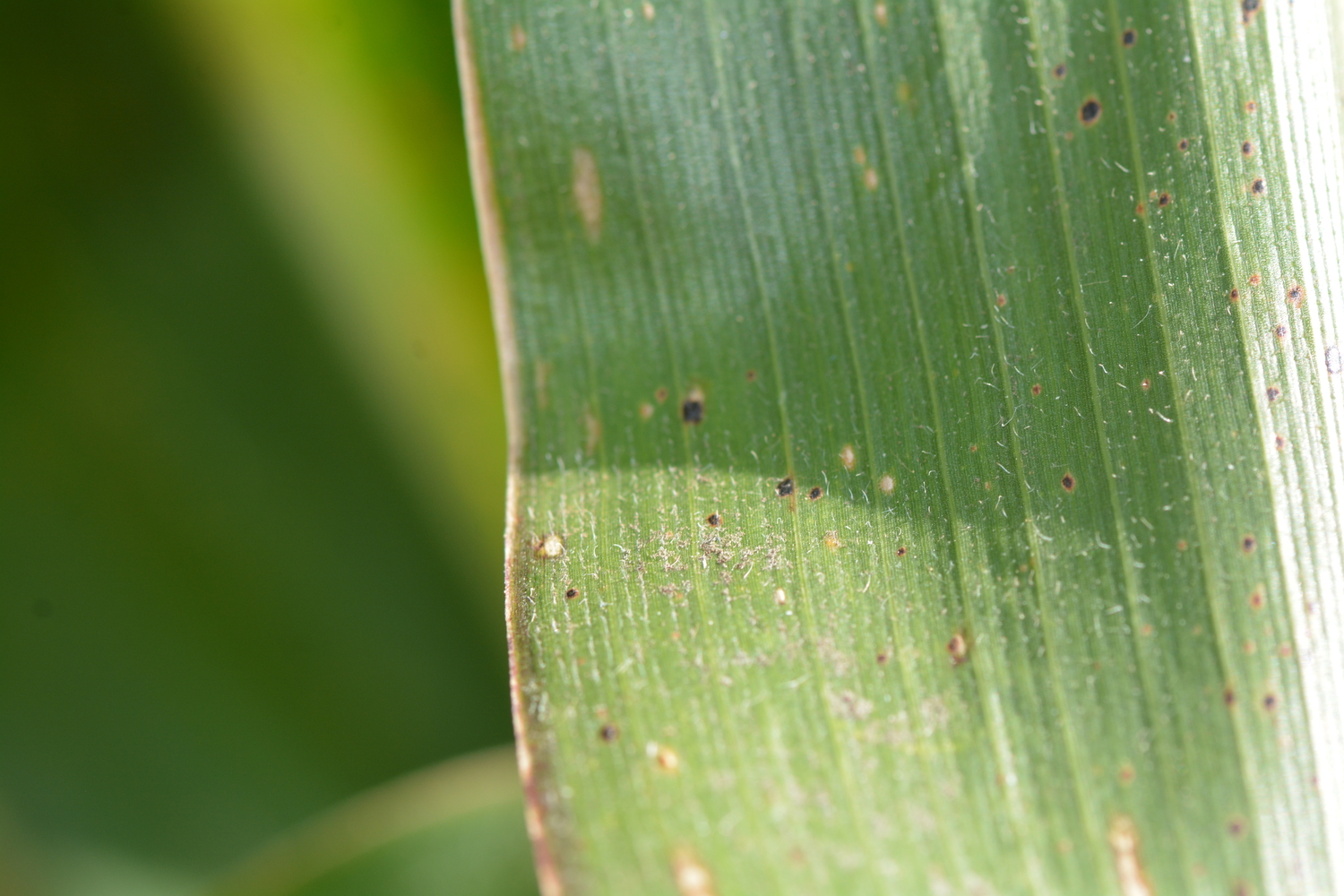}
       \includegraphics[width=\textwidth]
       {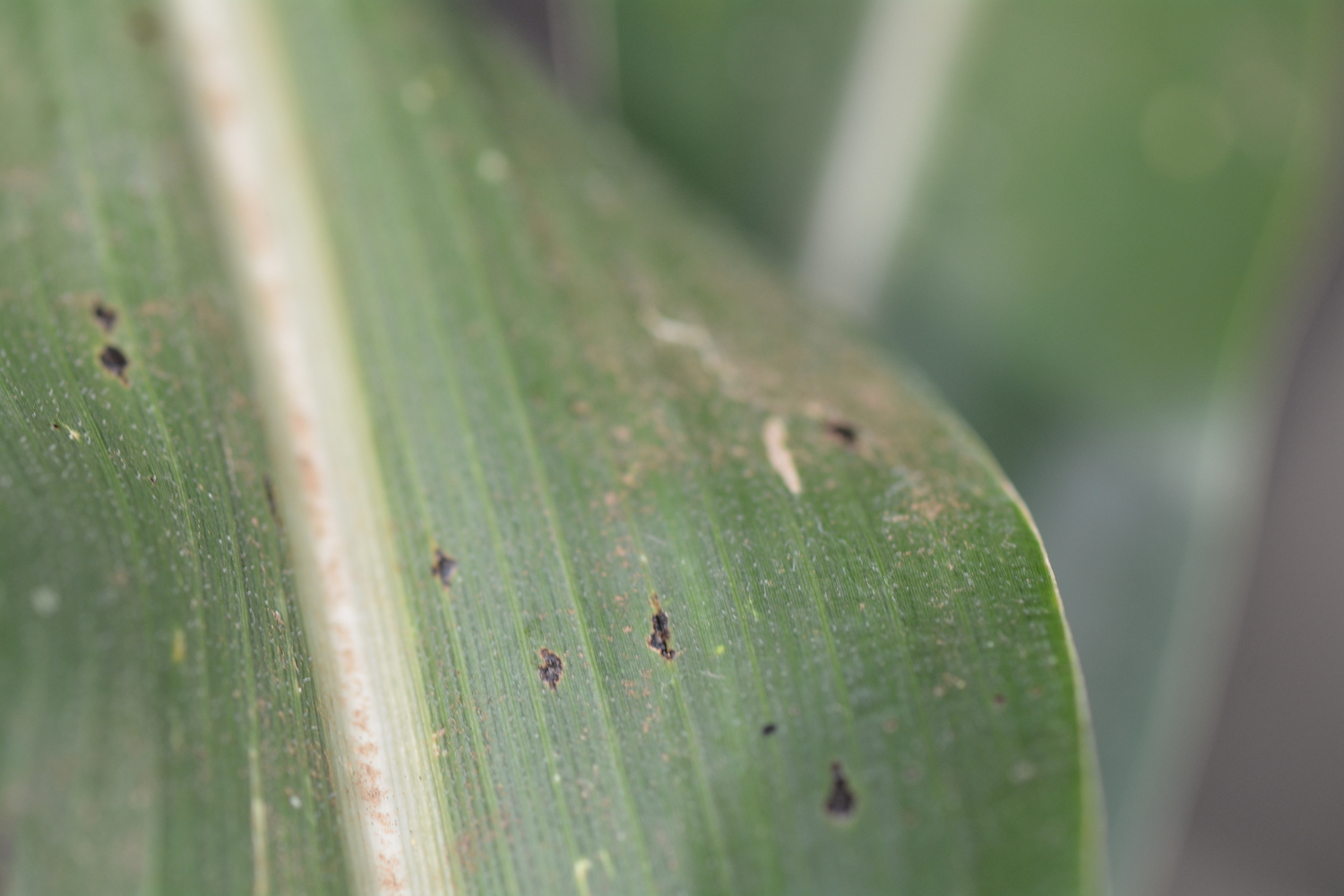}
       \caption{Close-up Image of Leaf with Tar Spot}
   \end{subfigure}
   \begin{subfigure}[t]{0.33\textwidth}
      \includegraphics[width=\textwidth]
      {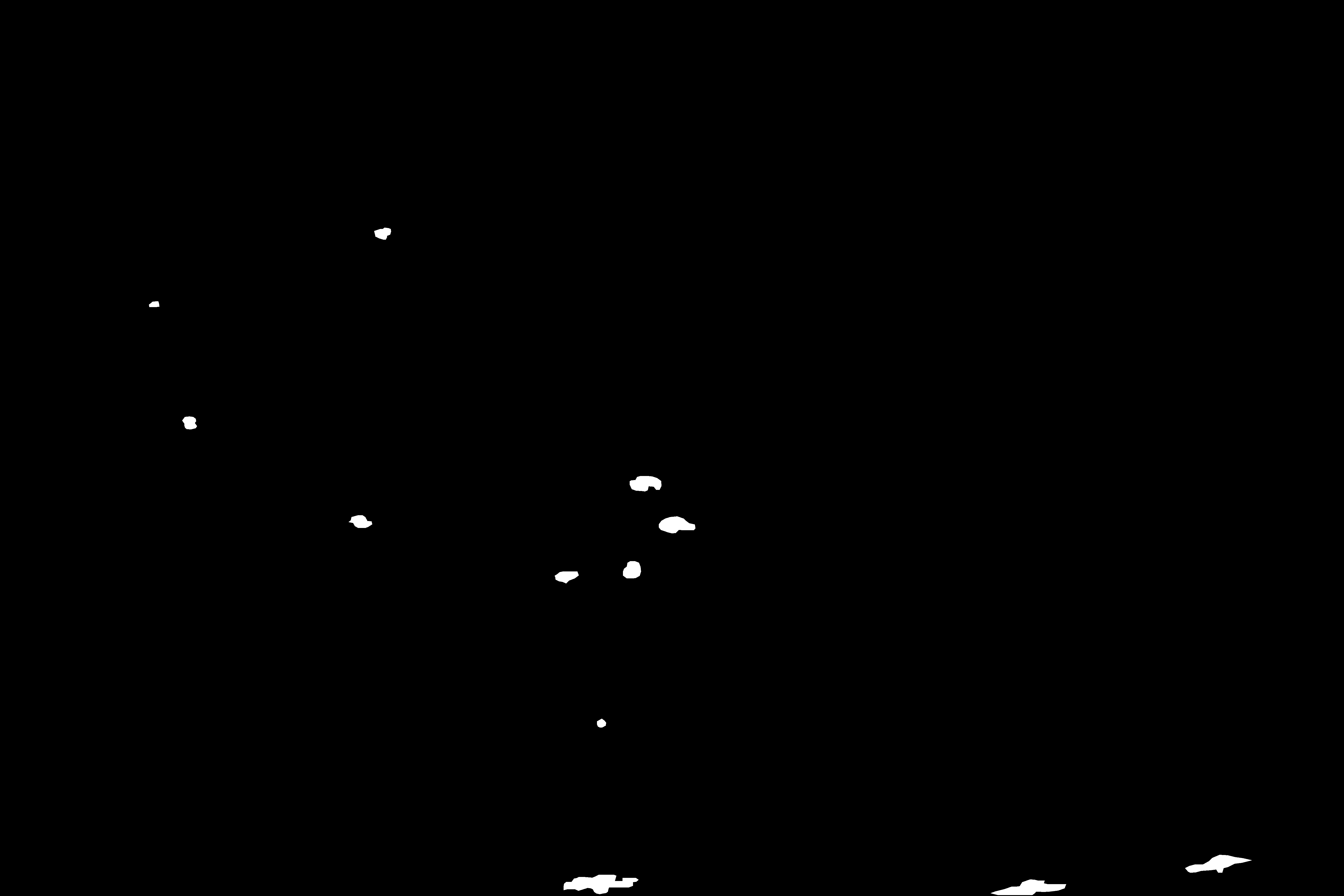}
      \includegraphics[width=\textwidth]
      {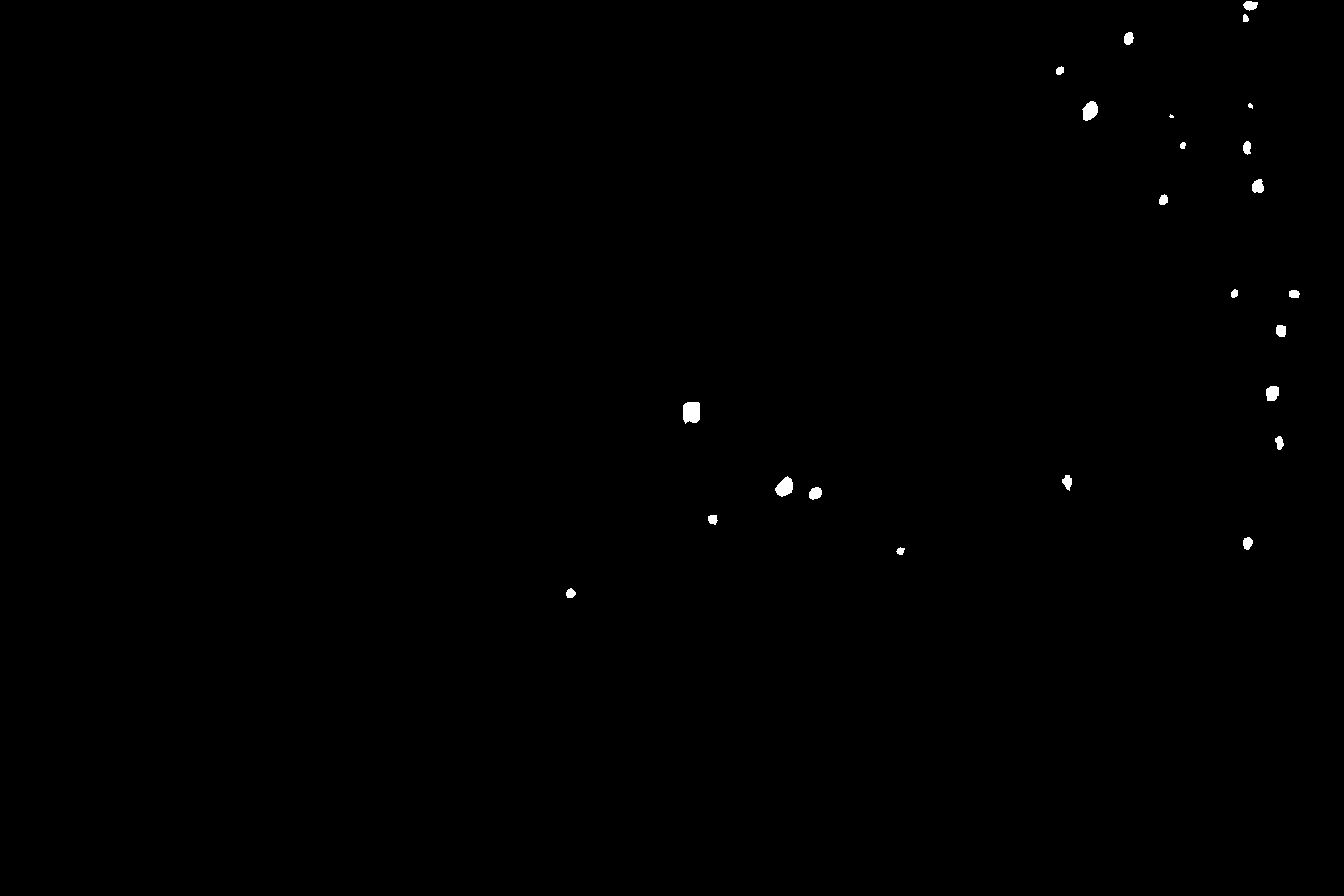}
      \includegraphics[width=\textwidth]
      {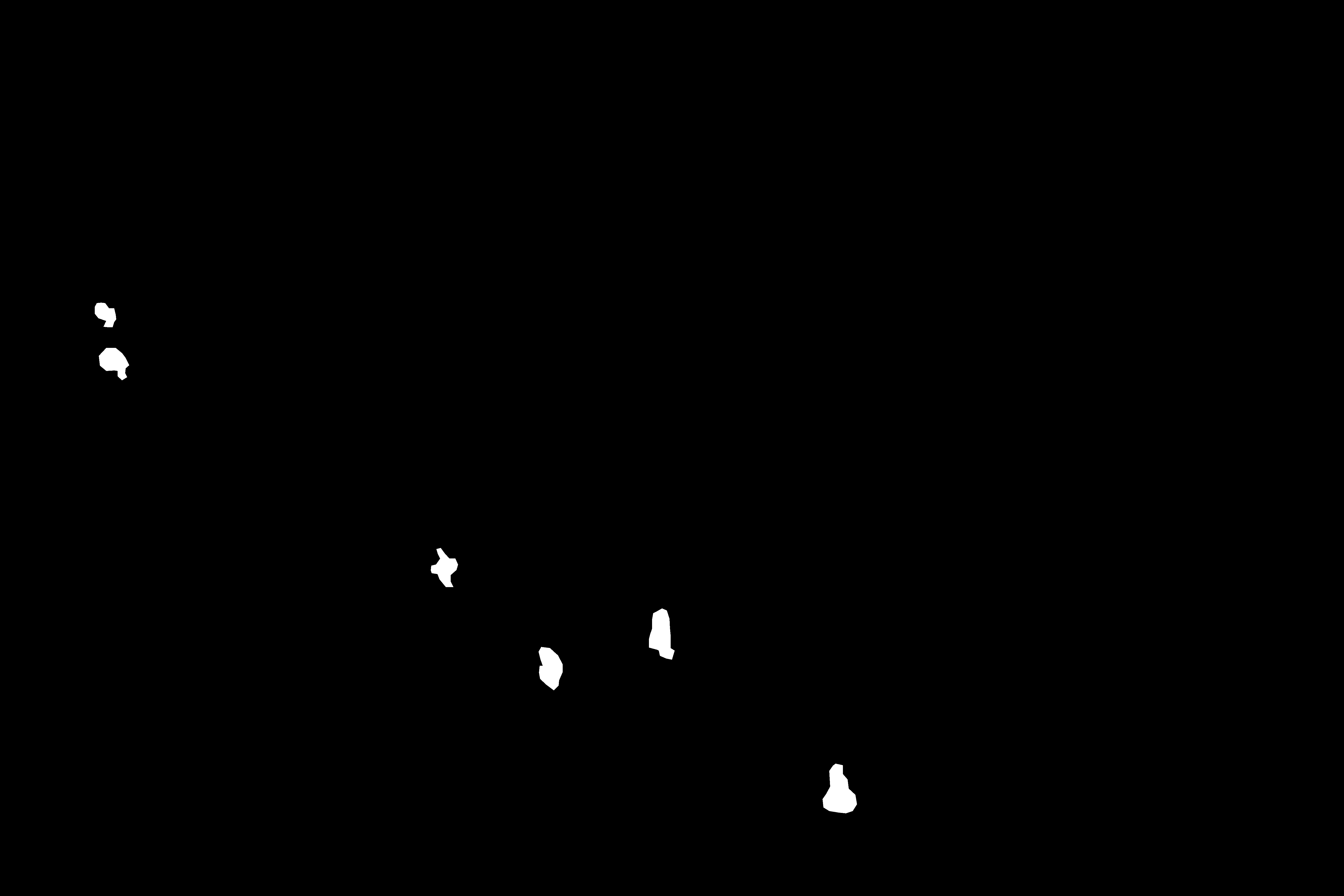}
      \caption{Tar Spot Ground Truth}
   \end{subfigure}
   \begin{subfigure}[t]{0.33\textwidth}
      \includegraphics[width=\textwidth]
      {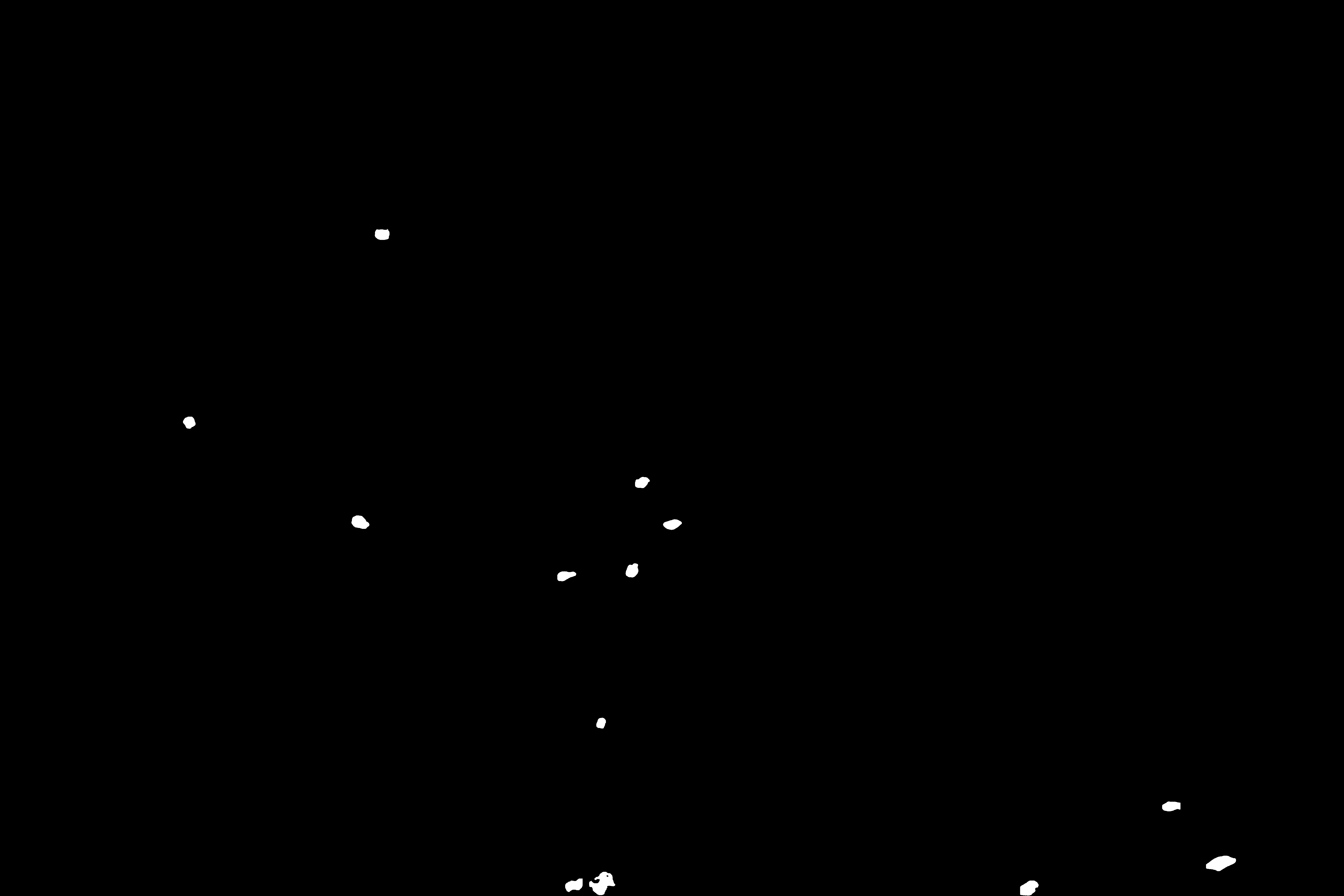}
      \includegraphics[width=\textwidth]
      {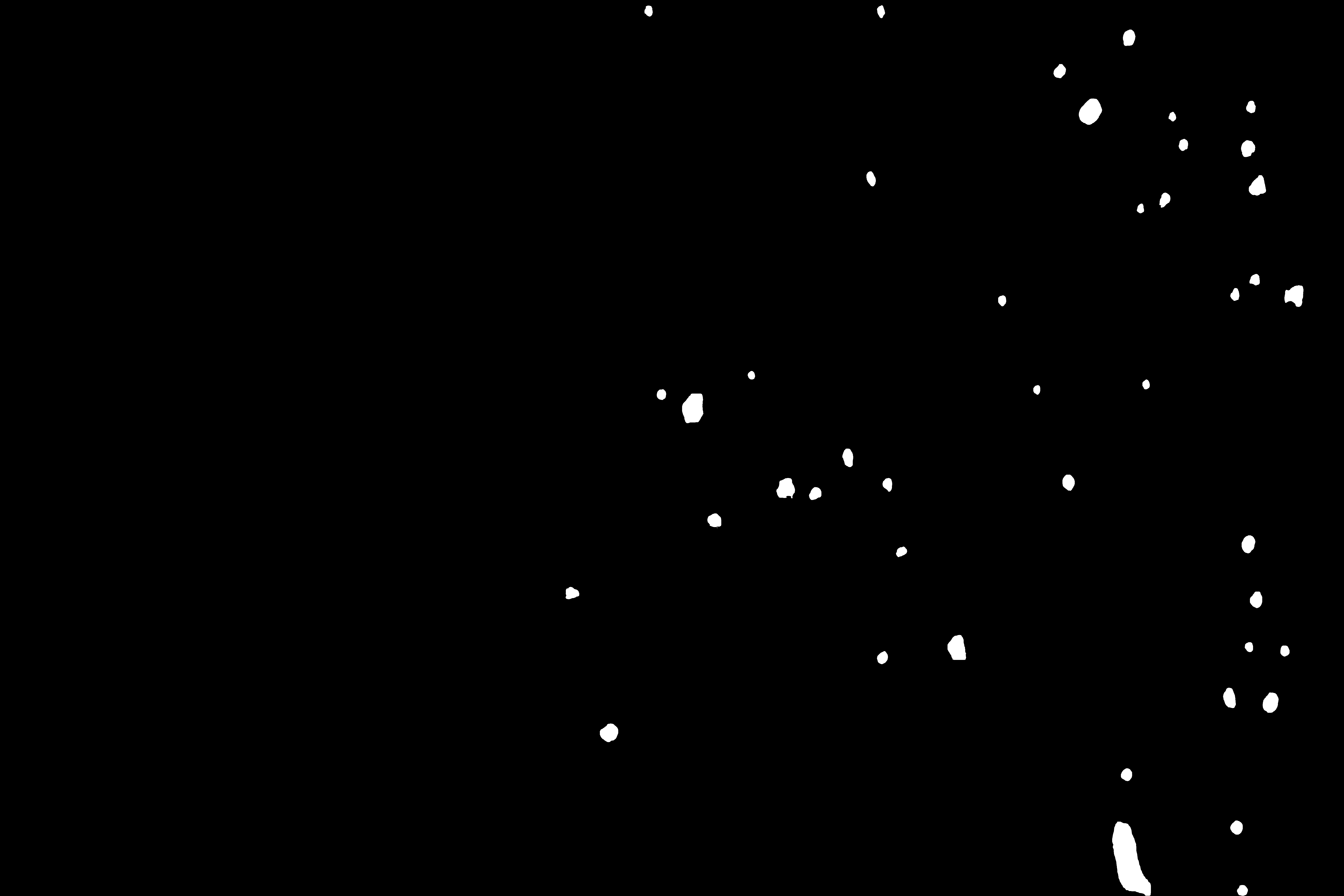}
      \includegraphics[width=\textwidth]
      {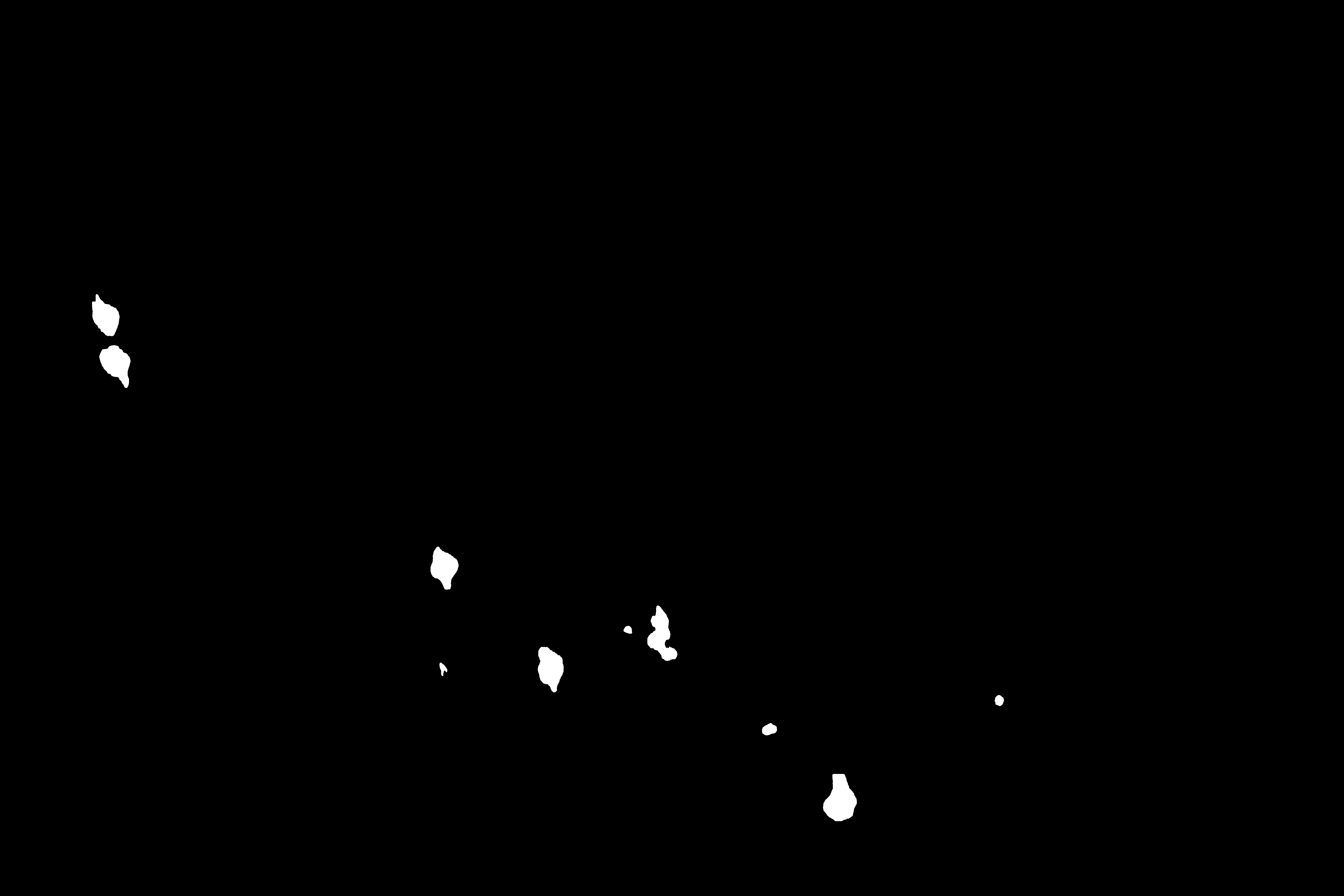}
      \caption{Tar Spot Detections}
   \end{subfigure}
   \caption{Tar spots detected by our trained Mask R-CNN. Note that the images also show the presence of plant disease symptoms other than tar spot, in addition to small debris often found on corn leaves in the field.}
   \label{fig:orig_results}
\end{figure*}

Our training dataset consists of $500$ close-up RGB images with a resolution of $6000 \times 4000$ pixels, captured by a Nikon D7100 camera.
The images are of leaves with tar spot, with between $0 - 150$ tar spots present in each image. 
We also have a manually labeled dataset consisting of $100$ similar images.
These images have been labeled by an expert plant pathologist, and as indicated previously, required a significant amount of time to ensure accuracy.
The labeled dataset is split into test and validation datasets by a ratio of $4:1$, meaning we have $20$ images to validate our trained model.
The validation dataset was also used to empirically select the best thresholds for our ``automatic ground truthing'' approach.
We generate the associated ground truth using our ``automatic ground truthing'' for the $500$ images in our training set.
We then train a Mask R-CNN with the 500 images until the validation loss no longer decreases.
Our trained Mask R-CNN achieves an F1-score of $0.76$ on the $80$ images in the manually ground truth testing dataset, as well as an average error of $10.4$ in counting the number of tar spots.
An example of the tar spots detected by the Mask R-CNN can be seen in Figure \ref{fig:orig_results}.

The other quantitative measurement of performance we are interested in is the average time taken to detect the tar spots in the leaf image.
We report these results in Table \ref{table:time_taken}, which show that using the Mask R-CNN leads to a drastic reduction in average time taken per image to generate tar spot detections.
Additionally, since we are leveraging the parallel processing and computational power of GPUs, it is comparatively easier to scale up and process thousands of leaf images from a field at once.
Thus, with this `automatic ground truthing' approach, we show that a deep learning model that facilitates high-throughput phenotyping can be developed.

\begin{table}[htbp]
   \centering
    \begin{tabular}{lc}
    \toprule
    Approach                    &Average Time \\
    \midrule
    Manual Annotation           &434 \\
    Automatic Ground Truthing   &26 \\
    Mask R-CNN                  &2 \\             
    \bottomrule
    \end{tabular}
    \caption{The average time (in seconds) to detect tar spots in a close-up leaf image.}
    \label{table:time_taken}
\end{table}

%\begin{figure*}[tbp]
%   \begin{center}
%   \includegraphics[width=0.7\linewidth]{figures/orig_output.JPG}
%   \end{center}
%      \caption{Minimal number of tar spots detected by the trained model on different type of data.}
%   \label{fig:zoomout_orig_results}
%\end{figure*}

%\begin{figure*}[tbp]
%   \begin{center}
%   \includegraphics[width=0.7\linewidth]{figures/merge_output.JPG}
%   \end{center}
%      \caption{Increased number of tar spots detected with the sliding window approach.}
%   \label{fig:zoomout_merge_results}
%\end{figure*}

\begin{figure*}[t]
   \centering
   \begin{subfigure}[t]{0.33\textwidth}
       \includegraphics[width=\textwidth]
       {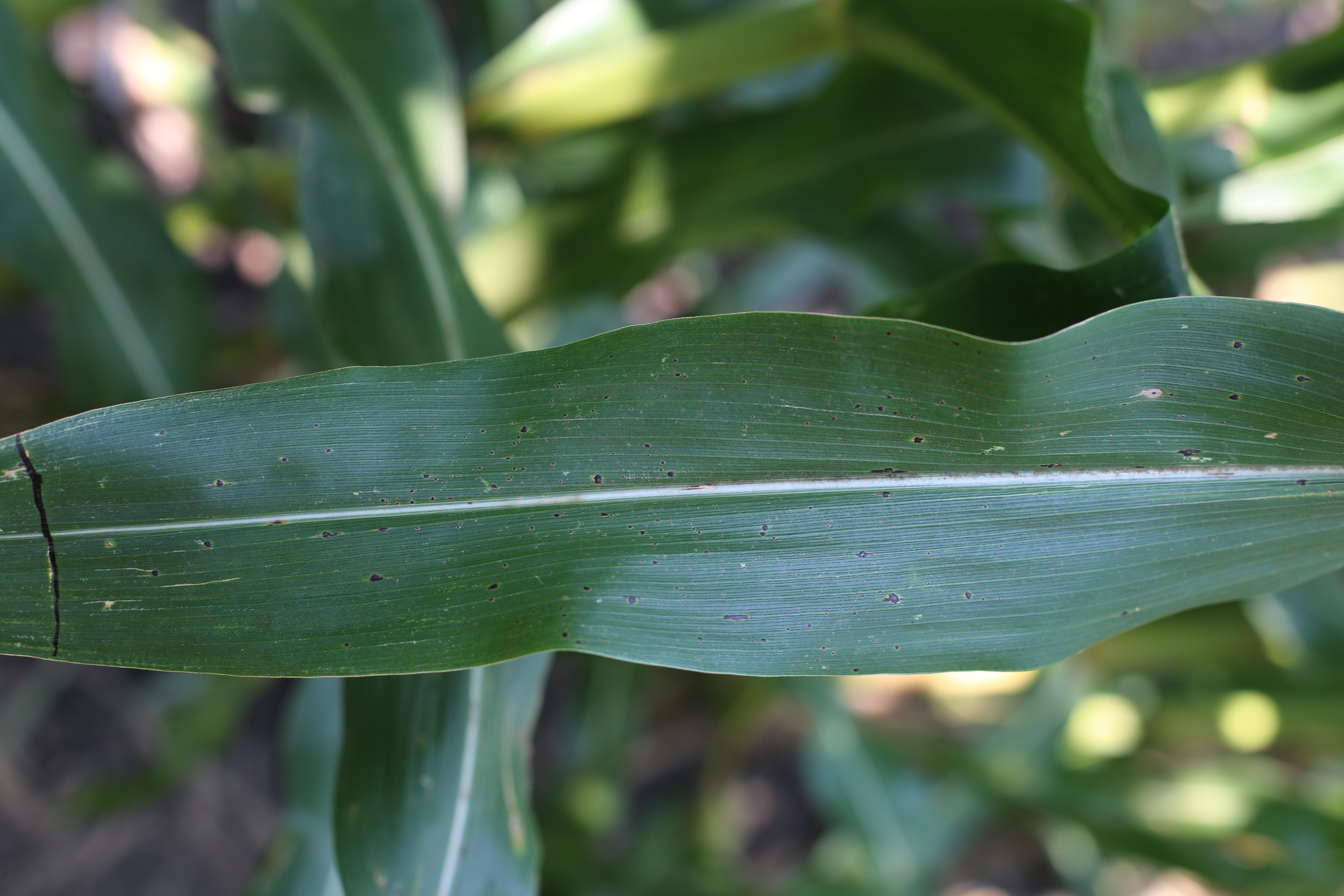}
       \includegraphics[width=\textwidth]
       {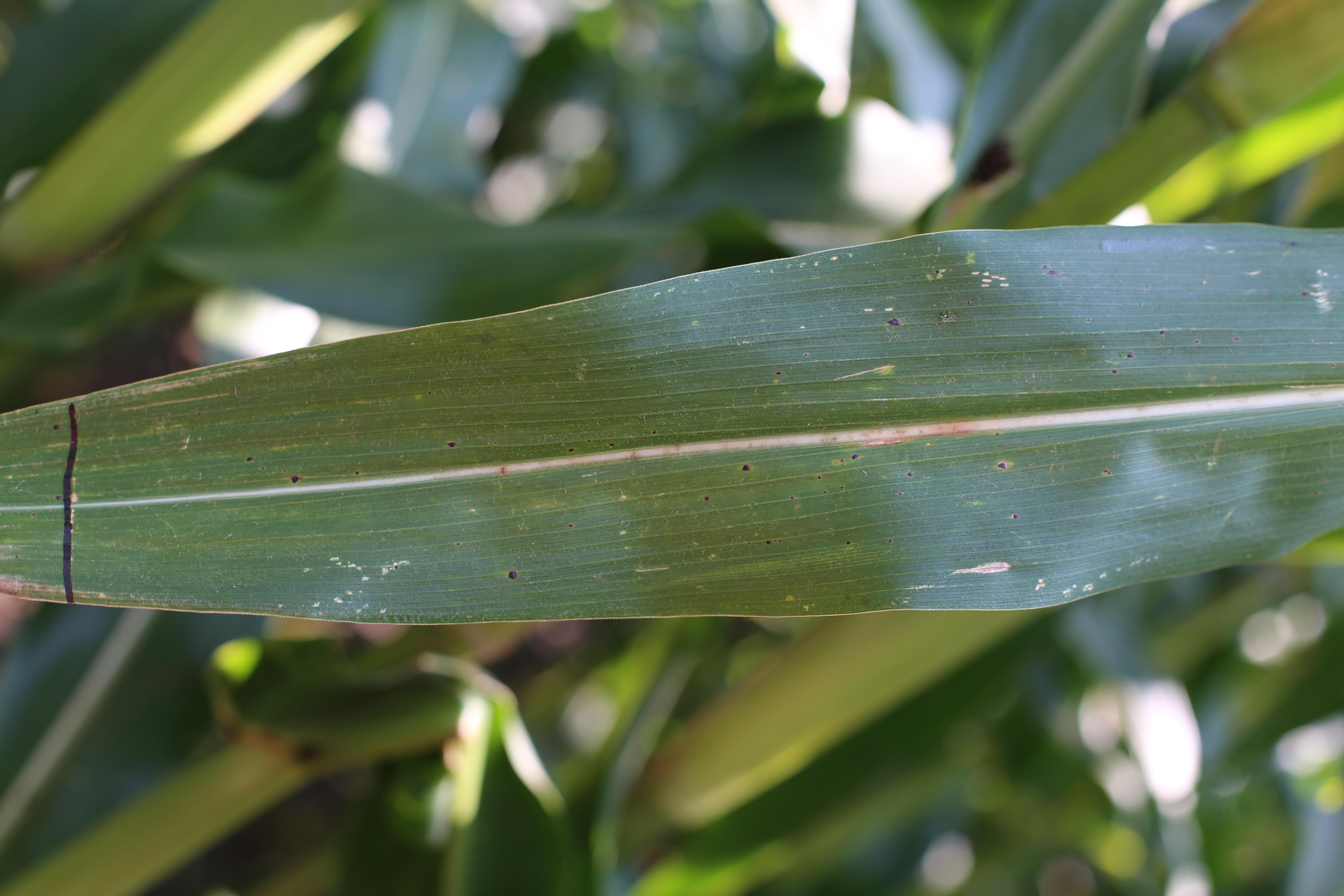}
       \includegraphics[width=\textwidth]
       {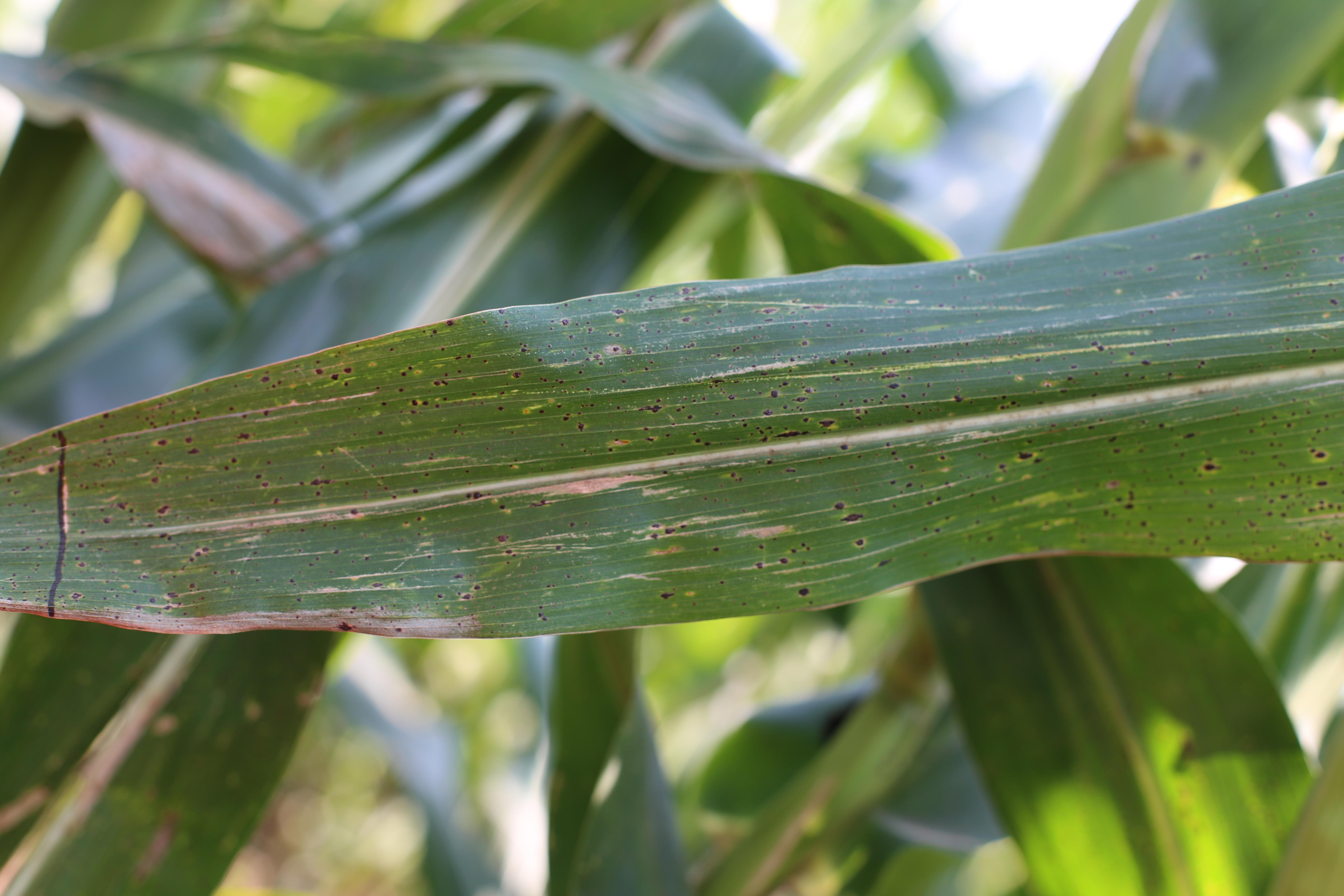}
       \caption{Full Image of Leaf with Tar Spot}
   \end{subfigure}
   \begin{subfigure}[t]{0.33\textwidth}
      \includegraphics[width=\textwidth]
      {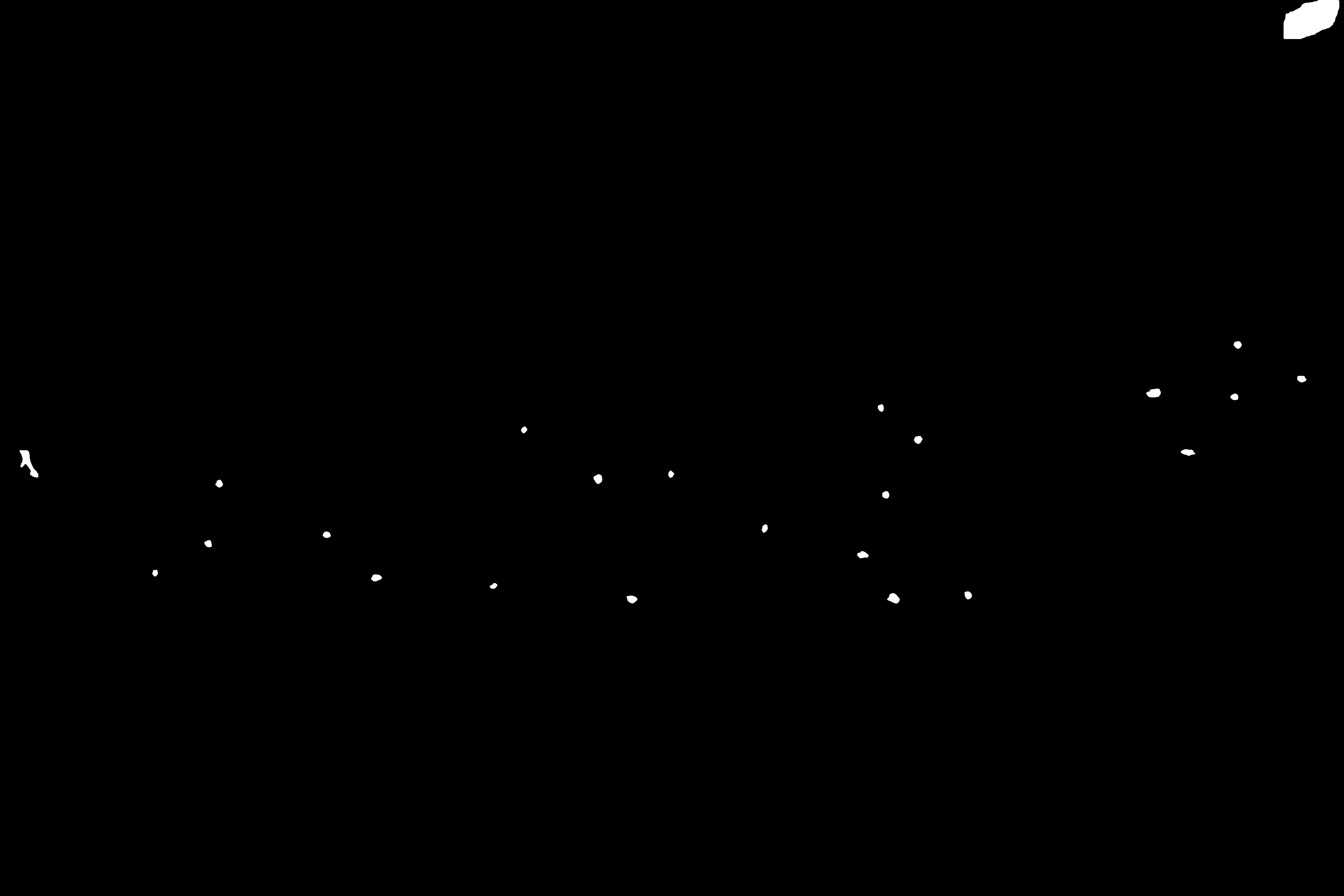}
      \includegraphics[width=\textwidth]
      {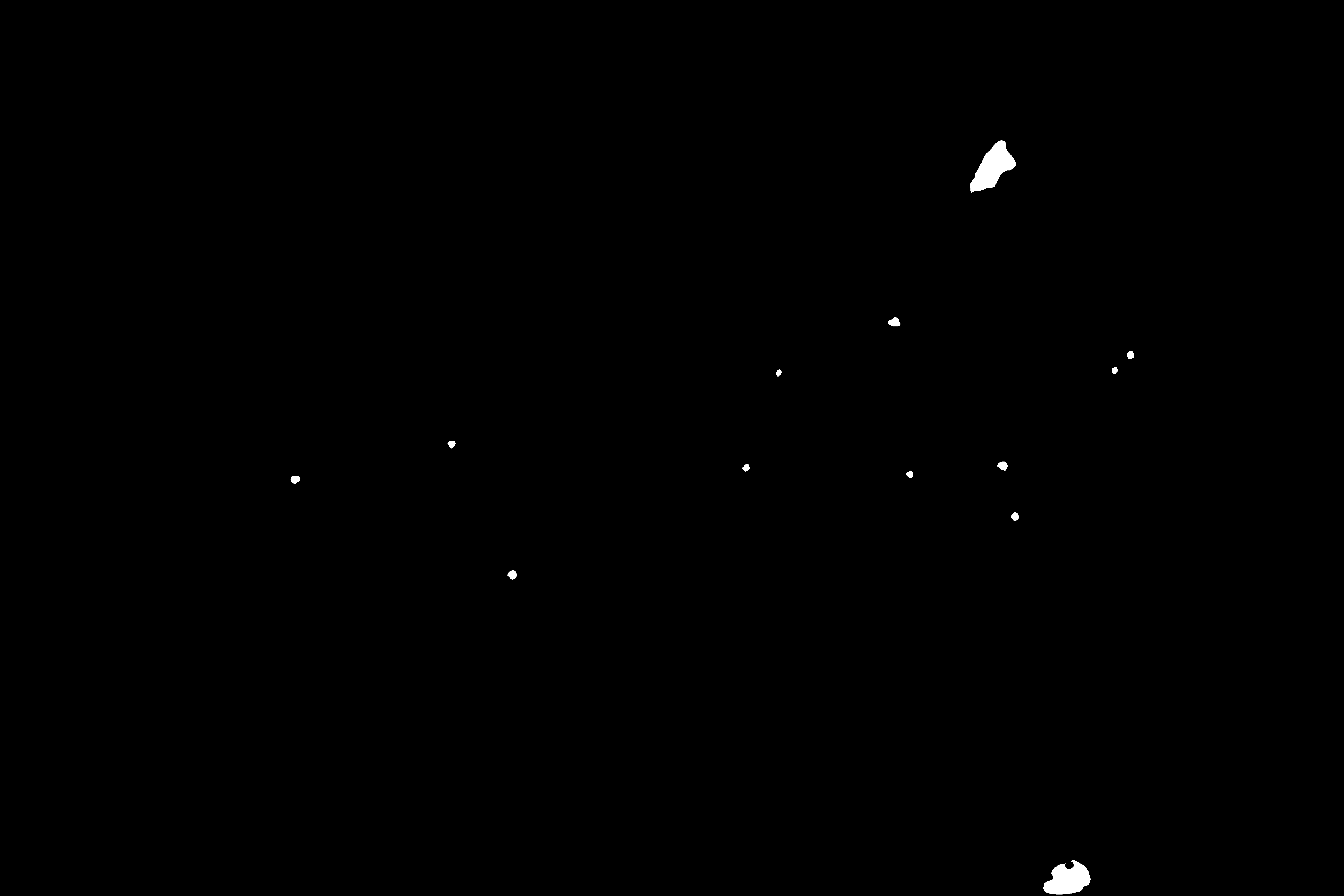}
      \includegraphics[width=\textwidth]
      {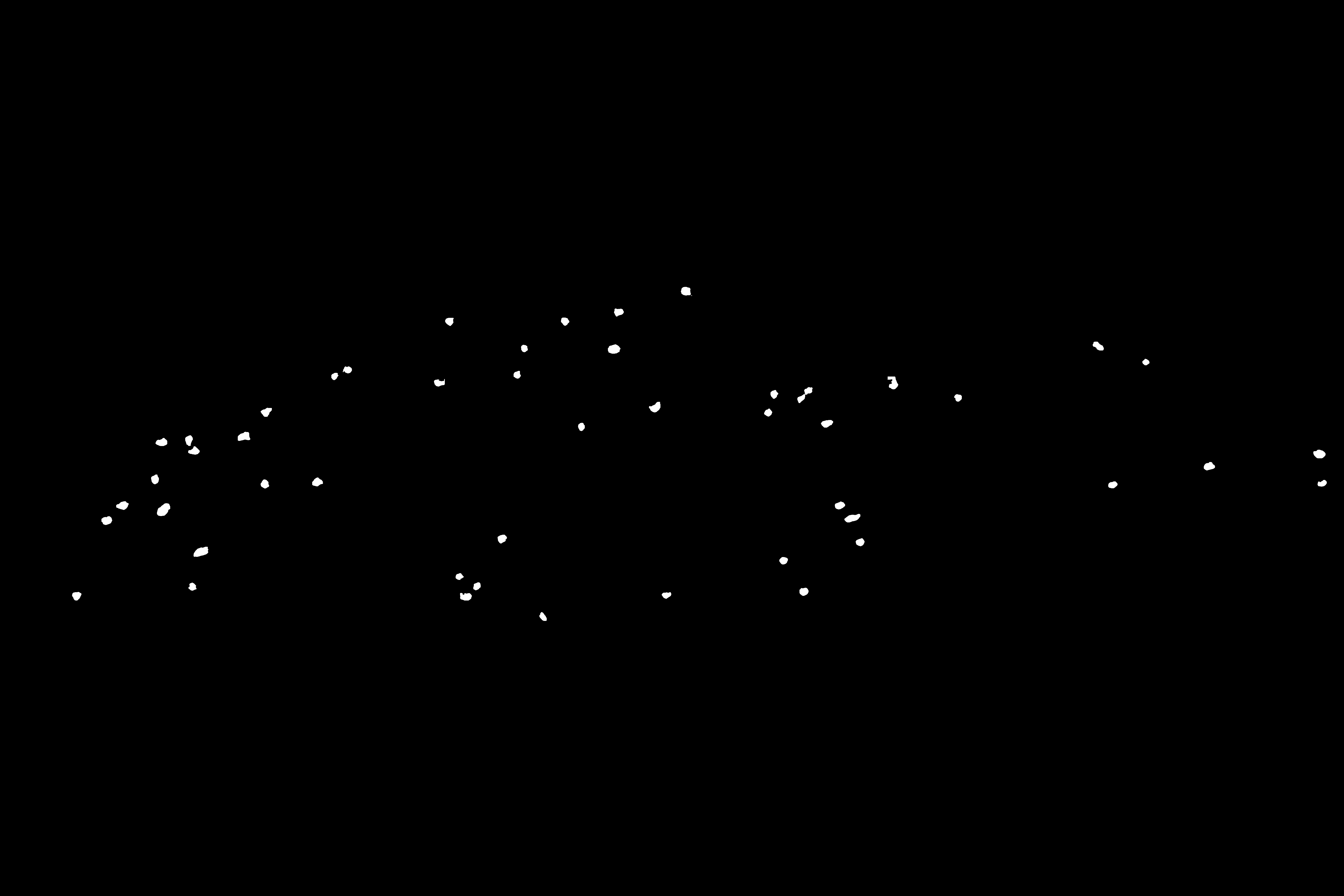}
      \caption{Initial Tar Spot Detections}
   \end{subfigure}
   \begin{subfigure}[t]{0.33\textwidth}
      \includegraphics[width=\textwidth]
      {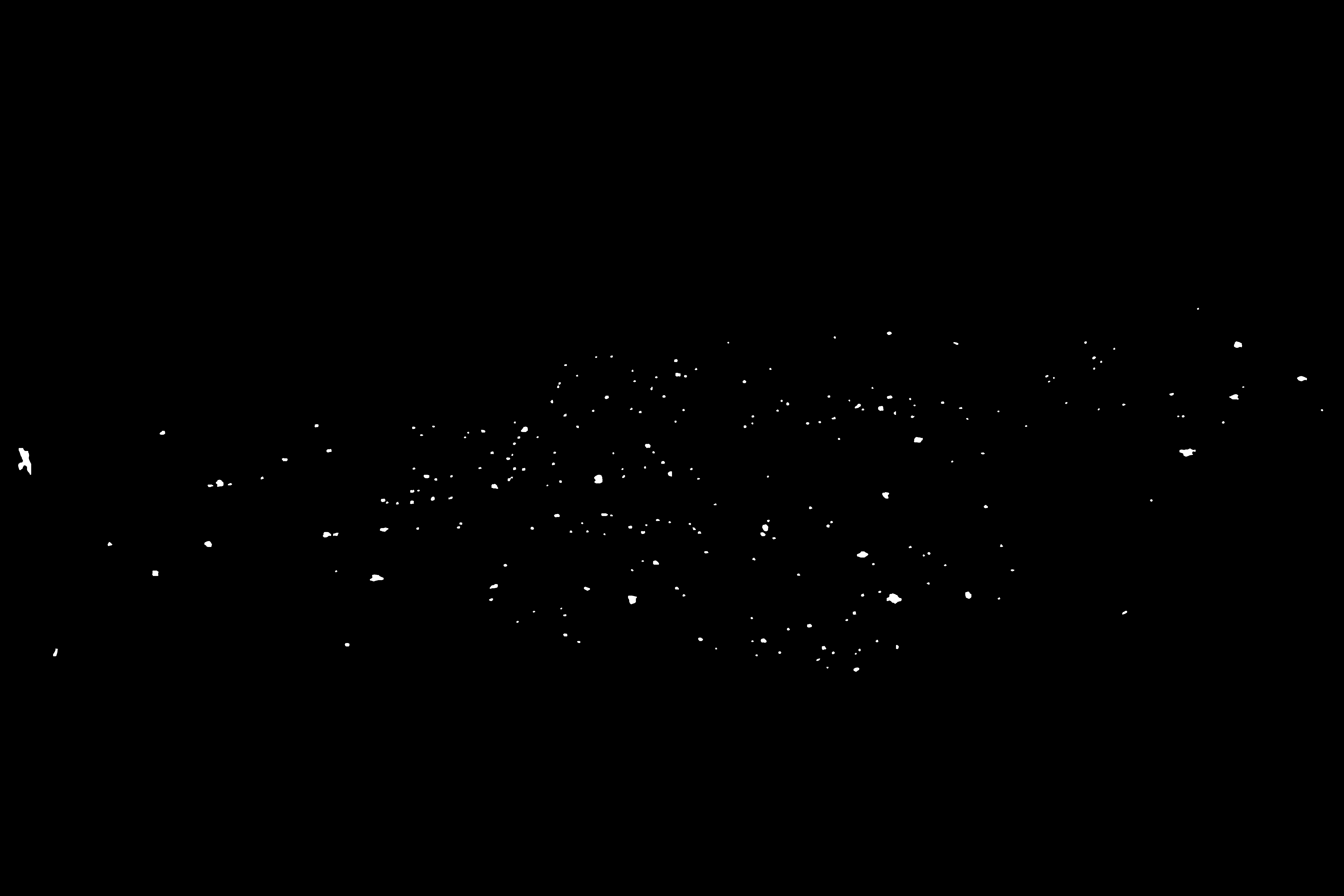}
      \includegraphics[width=\textwidth]
      {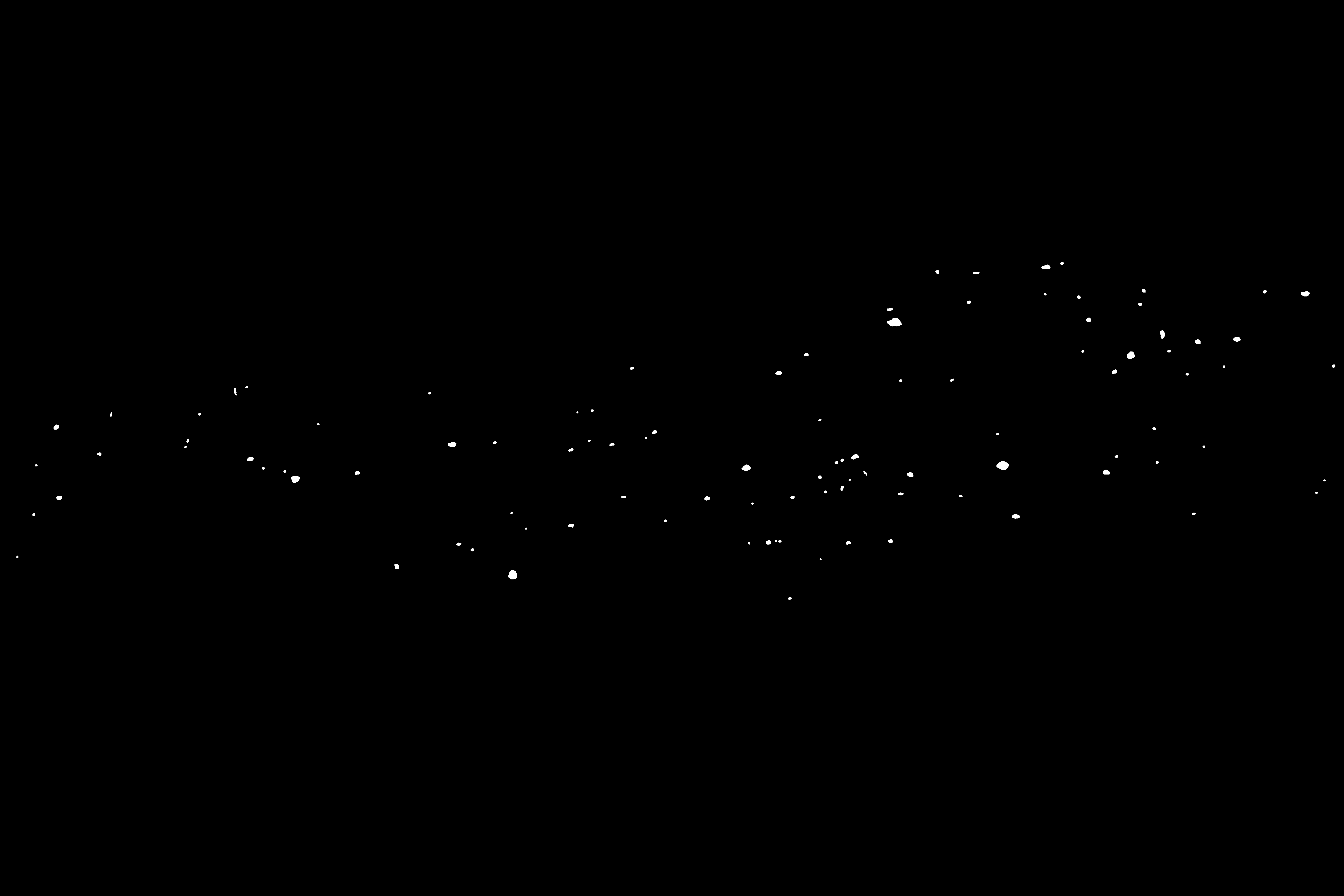}
      \includegraphics[width=\textwidth]
      {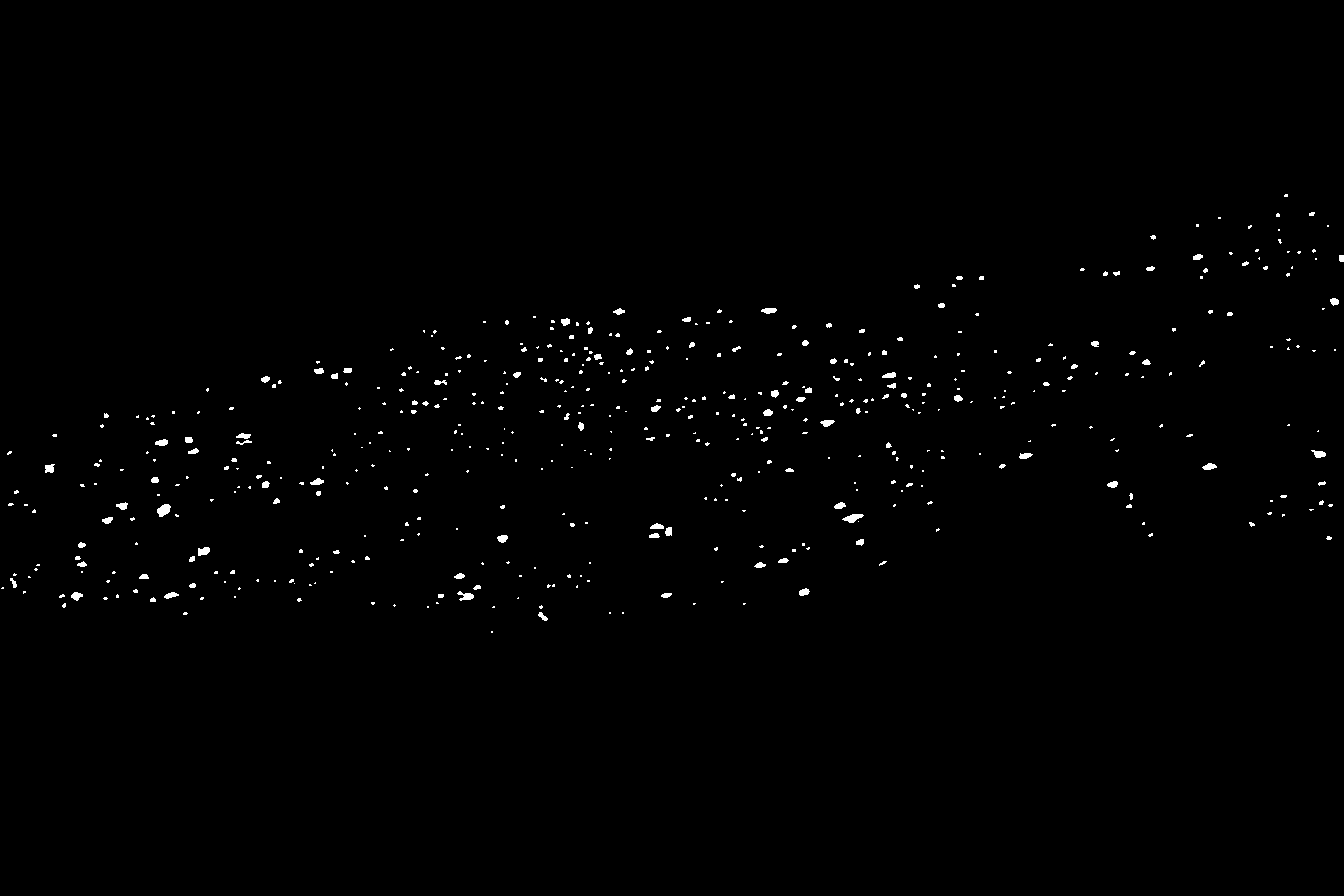}
      \caption{Sliding Window Tar Spot Detections}
   \end{subfigure}
   \caption{The full images of leaf with tar spot are difficult to analyze, manually or automatically. Our trained Mask R-CNN model captures a minimal number of tar spots, while adding the sliding window approach to the Mask R-CNN enables an increased number of the smaller tar spots to be detected.}
   \label{fig:zoomout_results}
\end{figure*}

The Mask R-CNN is trained to work on close-up images of leaves with tar spot, since these are the images where our threshold-based ground truthing approach works optimally, but also are the images that are the easiest to label for the expert. 
However, to obtain measurements that reflect the extent of the tar spot disease, the more practical approach is to acquire images from further away from the leaf that capture the leaf in its entirety. 
This ensures some consistency when measuring the relative area of the leaf that the tar spot disease has taken hold.
However, by moving further away, the tar spots in the image greatly reduce in size and increase in number. 
The second column of Figure \ref{fig:zoomout_results} shows the initial detection performance of our trained model on these types of images.

\begin{figure*}[tbp]
   \begin{center}
   \includegraphics[width=0.9\linewidth]{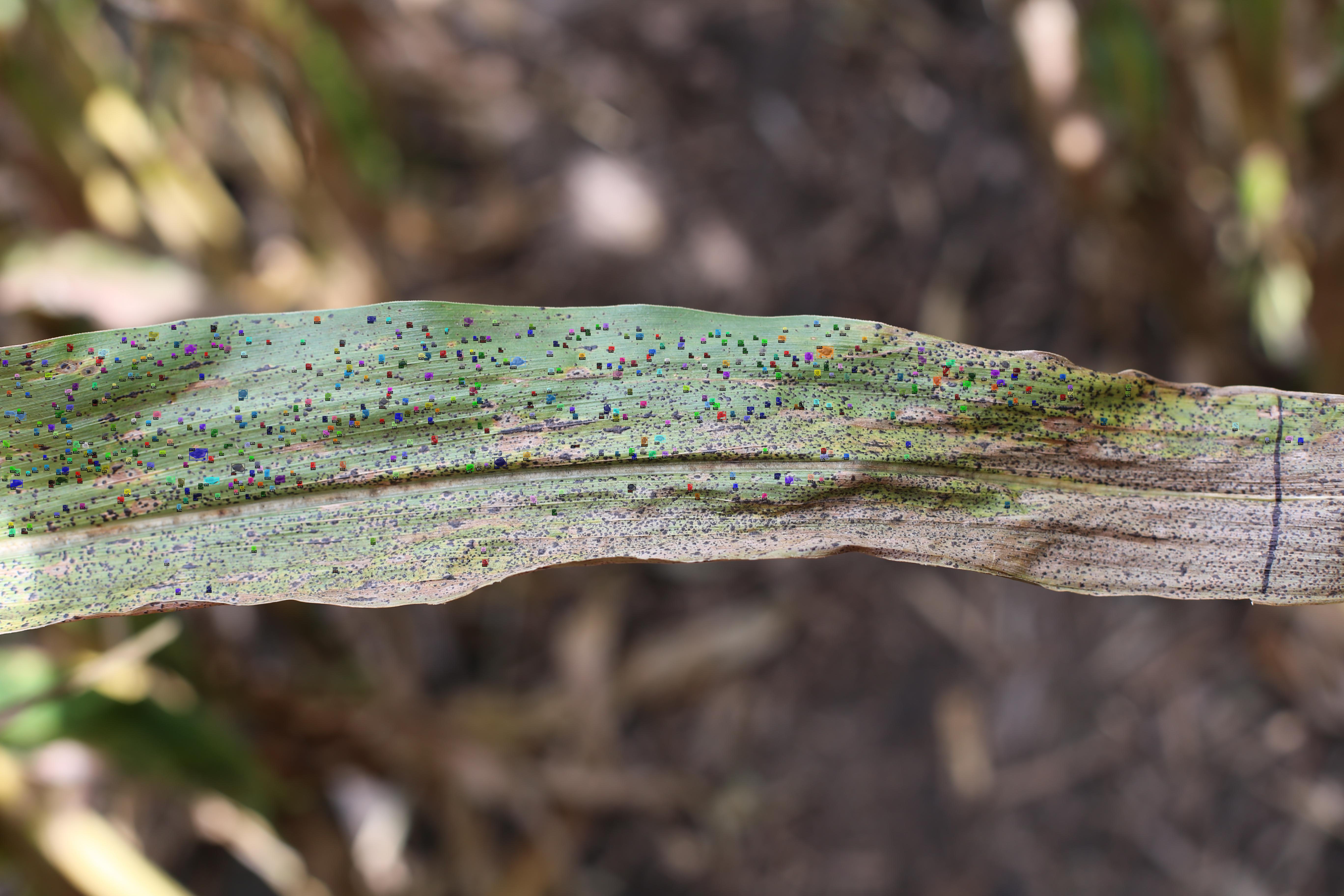}
   \end{center}
      \caption{An example of tar spot detections on a leaf with a very high tar spot density.}
   \label{fig:too_dense}
\end{figure*}

The larger tar spots are generally still identified, but a drawback of using a CNN-based approach is that smaller details in an image can be lost during the downsampling and feature extraction steps of the CNN. 
For these images, we make a slight modification to how we input the image to the network. 
We consider a sliding window of size $600 \times 400$. 
Its horizontal stride is $75$ and its vertical stride is $50$. 
We provide these $600 \times 400$ patches to our Mask R-CNN and detect tar spots for each patch. 
These patches are then combined using a voting strategy into a larger $6000 \times 4000$ image that reflects the tar spots detected for the entire image. 
The third column in Figure \ref{fig:zoomout_results} shows the results using this approach. 
Qualitatively, we can see that a greater number of tar spots are detected, and many of the smaller spots missed earlier are now being captured.

It is also important to note that the cost of manual annotation of these images increases exponentially. 
As can be seen in the examples, the number of tar spots increases tenfold and each one greatly reduces in size.
An expert would have to painstakingly examine every pixel in the image to completely capture the extent of the disease.

Even though this approach of using automatically generated ground truth worked reasonably well,
 there is still room for improvement when dealing with leaves with a high density of tar spots. 
The tar spots often begin to merge into irregular shapes or are too close to properly distinguish, especially when the whole leaf image is being used. 
An example of a high tar spot density leaf is shown in Figure \ref{fig:too_dense}. 
Although not every tar spot is detected, the relative relationship between images is maintained, i.e., we detect more tar spots in the leaf image with more tar spots, despite not capturing all of the tar spots. 
In terms of analysis, we are still generally able to capture the useful trend of increasing disease severity which is important for managing field based tar spot infections.

\section{Conclusions and Future Work}

In this paper, we show that a Mask R-CNN is very effective at detecting tar spots in corn leaf images.
Since manual annotation is very inefficient, we use image analysis techniques to automatically generate ground truth images for the Mask R-CNN.
With this approach, we are able to achieve reasonable tar spot detection and area estimation performance on close-up images of leaves in an automated manner, greatly reducing the average time spent to extract the relevant information from an image of a tar spot leaf.
We also show that this trained model can be applied with a sliding window approach to in-field images of whole leaves to capture the extent of disease progression in the field.
%\textbf{{\color{red} You need to say more about the performance.}}
%\textbf{{\color{blue} I have added some more.}}
%This enables us to leverage the parallel processing and computation power of GPUs to process many images quicker than traditional image processing.
Future work includes investigating the feasibility of semi-supervised learning i.e., using a limited amount of labeled data to leverage vast amounts of unlabeled data and train a neural network for a task.

%%%%%%%%% REFERENCES
{\small
\bibliographystyle{ieee_fullname}
\bibliography{egbib}
}

\end{document}